\documentclass[runningheads]{eccv2024/llncs}

\usepackage{eccv2024/eccv}

\usepackage{eccv2024/eccvabbrv}

\usepackage{graphicx}
\usepackage{booktabs}

\usepackage[accsupp]{axessibility}  

\usepackage{hyperref}

\usepackage{orcidlink}

\definecolor{cvprblue}{rgb}{0.21,0.49,0.74}
\usepackage[linesnumbered, ruled, noend]{algorithm2e}
\usepackage{xcolor}
\usepackage{multirow}
\usepackage{makecell}
\usepackage{wrapfig}
\usepackage{multicol}

\definecolor{behavior-encoder}{rgb}{0.341, 0.729, 0.69}
\definecolor{map-encoder}{rgb}{0.957, 0.792, 0.510}
\definecolor{init-encoder}{rgb}{0.396, 0.639, 0.831}
\definecolor{decoder}{rgb}{0.525, 0.514, 0.788}

\begin{document}

\title{RealGen: Retrieval Augmented Generation for Controllable Traffic Scenarios} 

\titlerunning{RealGen: Retrieval Augmented Generation for Controllable Traffic Scenarios}

\author{Wenhao Ding*\inst{1,2} \and
Yulong Cao*\inst{2} \and
Ding Zhao\inst{1} \and 
Chaowei Xiao\inst{2,3} \and
Marco Pavone\inst{2,4}
}

\authorrunning{W.~Ding et al.}

\institute{Carnegie Mellon University \and
NVIDIA Research\and
UW-Madison\and
Stanford University  \\
\email{\{wenhaod, yulongc\}@nvidia.com} 
}

\maketitle

\begin{abstract}
Simulation plays a crucial role in the development of autonomous vehicles (AVs) due to the potential risks associated with real-world testing. 
Although significant progress has been made in the visual aspects of simulators, generating complex behavior among agents remains a formidable challenge. 
It is not only imperative to ensure realism in the scenarios generated but also essential to incorporate preferences and conditions to facilitate controllable generation for AV training and evaluation. 
Traditional methods, which rely mainly on memorizing the distribution of training datasets, often fail to generate unseen scenarios. 
Inspired by the success of retrieval augmented generation in large language models, we present \textbf{RealGen}, a novel retrieval-based in-context learning framework for traffic scenario generation. 
RealGen synthesizes new scenarios by combining behaviors from multiple retrieved examples in a gradient-free way, which may originate from templates or tagged scenarios. 
This in-context learning framework endows versatile generative capabilities, including the ability to edit scenarios, compose various behaviors, and produce critical scenarios.
Evaluations show that RealGen offers considerable flexibility and controllability, marking a new direction in the field of controllable traffic scenario generation. Check our project website for more information: \href{https://realgen.github.io/}{https://realgen.github.io}.
\end{abstract}    
\section{Introduction}
\label{sec:intro}

Simulation is indispensable in the development of autonomous vehicles (AVs), primarily due to the considerable risks associated with training and evaluating these systems in real-world conditions. The biggest challenge in simulations lies in achieving realistic driving scenarios, as this realism influences the discrepancy between AV performance in simulated and actual environments. Although advancements in high-quality graphical engines have significantly enhanced the perception quality of simulators, the realism of agent behavior remains constrained because of the complicated interactions among naturalistic agents. To counteract this issue, data-driven simulation has emerged as a promising approach in the realm of autonomous driving, which leverages real-world scenario datasets to accurately generate the behaviors of agents.

With the rapid achievement of deep generative models~\cite{ho2020denoising} and imitation learning algorithms~\cite{hu2022model}, current data-driven simulations~\cite{li2023scenarionet, gulino2023waymax} can generate scenarios that closely mimic human driver behavior. 
However, the effectiveness of these simulations in accelerating the development of AV is limited. This limitation stems from the need for scenarios that meet specific conditions tailored for targeted training and evaluation. 
Achieving such controllability in simulations is challenging due to the complex nature of driving scenarios, which involve intricate interactions, diverse road layouts, and varying traffic regulations.

\begin{wrapfigure}{r}{0.6\textwidth}
    \centering
    \vspace{-8mm}
    \includegraphics[width=0.6\textwidth]{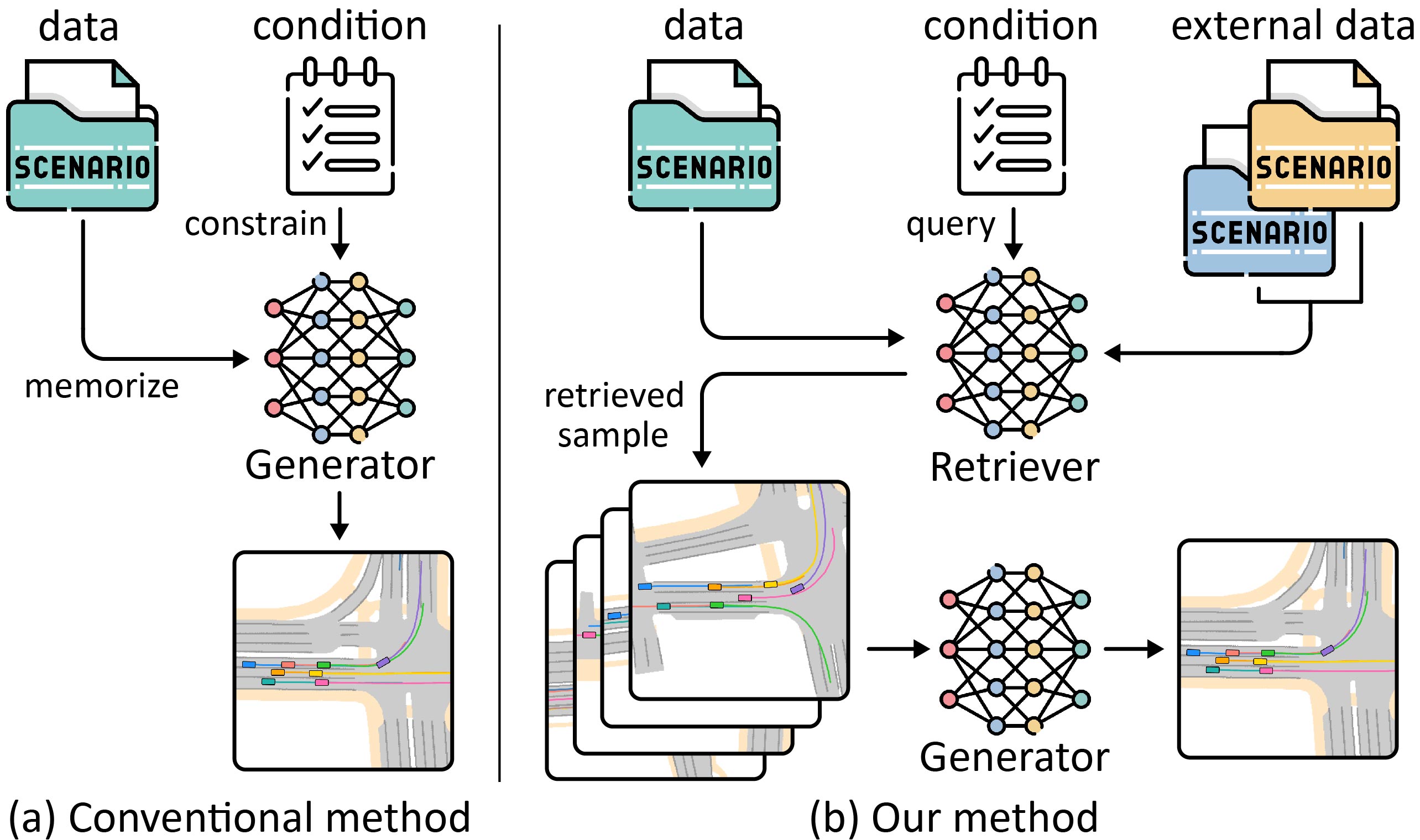}
    \vspace{-5mm}
    \caption{(a) Conventional methods make the model memorize the data distribution for generating. (b) In contrast, our method employs a retriever to query datasets (including external data obtained after training) and uses a generative model to generate scenarios by integrating the information from the retrieved scenarios.}
    \label{fig:intro}
    \vspace{-7mm}
\end{wrapfigure}


In pursuit of controllability, existing work applies additional guidance, typically in the form of constraint functions~\cite{ding2021semantically, zhong2023guided} or languages~\cite{tan2023language, zhong2023language}, to pre-trained scenario generative models. 
Regularization of the generation process through these tools is straightforward and effective, yet it encounters two principal challenges.
First, the training of scenario generative models typically utilizes naturalistic datasets, which might not encompass the specific scenarios desired as per the control signals. Even if such scenarios exist within the dataset, they are often omitted because of the rarity of long-tail data. 
The second challenge is that the representation of the guidance to the generative model may not be sufficiently expressive to accurately depict complex scenarios, such as specifying intricate interactions among multiple vehicles, using language. These limitations underscore the need for a more sophisticated and nuanced framework for controllable scenario generation.

Retrieval Augmented Generation (RAG)~\cite{chen2017reading, guu2020retrieval}, which enhances the generative process by querying related information from external databases, represents great potential in the domain of large language models~\cite{lewis2020retrieval}. 
In contrast to conventional models that memorize all knowledge within their parameters, as shown in Figure~\ref{fig:intro}(a), RAG models, shown in Figure~\ref{fig:intro}(b), learn to generate comprehensive outputs by retrieving pertinent knowledge from a database, based on the input provided. 
A notable aspect of RAG is the ability of the database to undergo updates even after the model has been trained, allowing continuous improvement and adaptation.
This flexible framework offers possibilities for controllable scenario generation by using appropriate template scenarios as input and facilitating the generation that is not only realistic but also aligned with specific training and evaluation requirements.

In this work, we present \textit{RealGen}, a retrieval augmented generation framework for generating traffic scenarios.
This framework, as shown in Figure~\ref{fig:structure}, begins with the training of an encoder through contrastive self-supervised learning~\cite{oord2018representation} to allow the retrieval process to query similar scenarios in a latent embedding space. Leveraging this latent representation, we subsequently train a generative model that combines retrieved scenarios to create novel scenarios. The key contributions of this paper are summarized below.
\begin{itemize}
    \item We develop a novel contrastive autoencoder model to extract scenario embeddings as latent representations, which can be used for a wide range of downstream tasks.
    \item We propose the first retrieval augmented generation framework using the latent representation tailored for controllable driving scenario generation. 
    \item We validate our framework through qualitative and quantitative metrics, demonstrating strong flexibility and controllability of generated scenarios.
\end{itemize}

\section{Related Work}
\label{sec:related-work}

\subsection{Traffic Modeling and Scenario Generation}

Research on traffic modeling using generative models has received considerable attention in recent works. Notably, ScenarioNet~\cite{li2023scenarionet} and Waymax~\cite{gulino2023waymax} employ large-scale data to train imitation learning models, facilitating the generation of realistic multi-agent scenarios in simulations. SceneGen~\cite{tan2021scenegen} utilizes the Long Short Term Memory~\cite{hochreiter1997long} module to autoregressively generate the trajectories for vehicles and pedestrians based on provided maps. Additionally, TrafficSim~\cite{suo2021trafficsim} learns the multi-agent behaviors from real-world data, and TrafficGen~\cite{feng2023trafficgen} proposes a transformer-based autoencoder architecture to accurately model complex interactions involving multiple agents.
MixSim~\cite{suo2023mixsim} builds a reactive digital twin and finds safety-critical scenarios with black-box optimization.
RTR~\cite{zhang2023learning} models reactive behaviors of vehicles using the combination of reinforcement learning and imitation learning.

Numerous prior studies have employed adversarial generation to synthesize rare yet critical scenarios.
L2C~\cite{ding2020learning} utilizes a reinforcement learning framework, where the reward for scenario generation is based on the collision rate. 
To enhance the realism in purely adversarial generation methods, MMG~\cite{ding2021multimodal} incorporates the data distribution as regularization. Further developments such as AdvSim~\cite{wang2021advsim}, AdvDO~\cite{cao2022advdo}, and KING~\cite{hanselmann2022king} integrate vehicle dynamics to directly optimize the trajectory to find critical scenarios. 

The proliferation of large language models (LLMs) has led to recent approaches in generating traffic scenarios with language as conditions to follow instructions from humans. CTG++~\cite{zhong2023language} replaces the gradient guidance process of CTG~\cite{zhong2023guided} with cost functions generated by an LLM. Leveraging the power of GPT-4~\cite{yang2023dawn}, this framework can generate diverse motion data with language conditions. Additionally, LCTGen~\cite{tan2023language} also harnesses the strengths of GPT-4 to generate a heuristic intermediate representation of scenarios from language inputs. Subsequently, a generative model pre-trained on open-source datasets generates various scenarios with this representation as inputs.

\subsection{Retrieval-augmented Generation}

Information retrieval (IR)~\cite{ibrihich2022review} is the procedure of representing and searching a collection of data with the goal of extracting knowledge to satisfy the query of users. Predominantly utilized in search engines and digital libraries~\cite{roshdi2015information}, IR primarily deals with the information stored in textual formats. Recently, IR has extended across diverse sectors, notably enhancing the quality of outputs in areas such as language modeling~\cite{borgeaud2022improving, liu2022relational}, question answering~\cite{guu2020retrieval, zhang2021greaselm}, image creation~\cite{blattmann2022retrieval, chen2022re}, and the generation of molecular~\cite{wang2022retrieval}.

One intuitive usage of the retrieved information is to enhance input data through several methods, such as merging the original data and retrieved data~\cite{lewis2020retrieval}, employing attention mechanism~\cite{borgeaud2022improving}, or extracting skeleton~\cite{cai2019retrieval}. 
The rationale behind retrieval systems stems from the impracticality of encoding all knowledge within model parameters, especially considering the dynamic nature of knowledge that evolves with human activities.
Consequently, the capability to access external knowledge databases can significantly improve the precision and quality of generated responses, as evidenced in applications involving LLMs~\cite{lewis2020retrieval, borgeaud2022improving}.

Another usage of retrieved data, particularly pertinent to this study, involves the controllable generation by integrating desired features retrieved from the dataset.
In~\cite{kim2020retrieval}, the authors generate product reviews with controllable information about the user, product, and rating.
Similarly, the process of story creation can be viewed as a blend of external story databases with selected text fragments~\cite{xu2020megatron}
Furthermore, \cite{wang2022retrieval} explores the controllable generation of molecules to meet various constraints, utilizing a database comprising simple molecules that typically meet only one of these constraints.

\subsection{Self-supervised Feature Learning}

The necessity for manual labeling in supervised learning often results in human biases, extraneous noises, and labor-intensive efforts. Therefore, the paradigm of self-supervised learning (SSL)~\cite{shurrab2022self}  is garnering heightened interest, particularly for its promising applications in language modeling and image interpretation. SSL algorithms usually learn implicit representations from extensive pools of unlabeled data without relying on human annotations. Generally, this line of research can be bifurcated into two categories~\cite{zhang2022survey}: generative SSL and discriminative SSL.

In the domain of generative SSL, models employ an autoencoder to convert input data into a latent representation, followed by a reconstruction process.
As an illustrative example, Denoising Autoencoder~\cite{vincent2008extracting}, an early example of this approach, reconstructs images from their noisy version to extract representative features.
Another technique in this field is masked modeling, extensively utilized in language models such as GPT-3~\cite{brown2020language} and BERT~\cite{devlin2018bert}, which predicts the masked token to gain a semantic understanding of the text.
More recently, the Masked Autoencoder~\cite{he2022masked} has emerged as a powerful framework, demonstrating its efficacy across a variety of downstream tasks.

Discriminative SSL typically optimizes a discriminative loss to learn representations from sets of anchor, positive, and negative samples.
Without ground-truth labels, these pairs are often constructed through solving jigsaw puzzles~\cite{noroozi2016unsupervised} or making geometry-based prediction~\cite{gidaris2018unsupervised}.
A prominent example of discriminative SSL is contrastive learning, which brings samples from the same class closer while distancing those from different classes.
Representative works such as MoCo~\cite{he2020momentum} and SimCLR~\cite{chen2020simple} learn embeddings by capturing invariant features between original data and its augmented variants. 
Additionally, InfoNCE~\cite{oord2018representation}, a method grounded in noise contrastive estimation~\cite{gutmann2010noise}, is widely used due to its effectiveness.

\section{A Latent Representation of Scenario}
\label{sec:embedding}

A crucial element within the retrieval system is the selection metric for data retrieval, which is typically implemented through a distance function between the query sample and the candidate samples in the database. Unlike text, which can be converted into word embeddings, traffic scenarios encompass sequential behaviors and intricate interactions among entities, complicating the establishment of a similarity metric for these scenarios.
Consequently, in this section, we introduce a scenario autoencoder to extract latent representations that facilitate the assessment of similarity between various traffic scenarios, a vital component in our RAG framework.

\subsection{Scenario Definition}

Each scenario is characterized by the trajectories $\tau \in \mathbb{R}^{M\times T \times 5}$, encompassing $M$ agents over a maximum of $T$ time steps. The trajectory of each agent is composed of parameters $[x, y, v, c, s]$, signifying position $x$, position $y$, velocity $v$, cosine of the heading $c$, and sine of the heading $s$. The initial state of these entities is denoted as $\tau_0 \in \mathbb{R}^{M \times 5}$.
Furthermore, the map is encapsulated by $m \in \mathbb{R}^{S\times 4}$, composed of $S$ lane segments. 
The attributes of these points $[x_s, y_s, x_e, y_e]$ correspond to the starting and ending positions of each segment.

\begin{figure*}[t]
    \centering
    \includegraphics[width=1.0\textwidth]{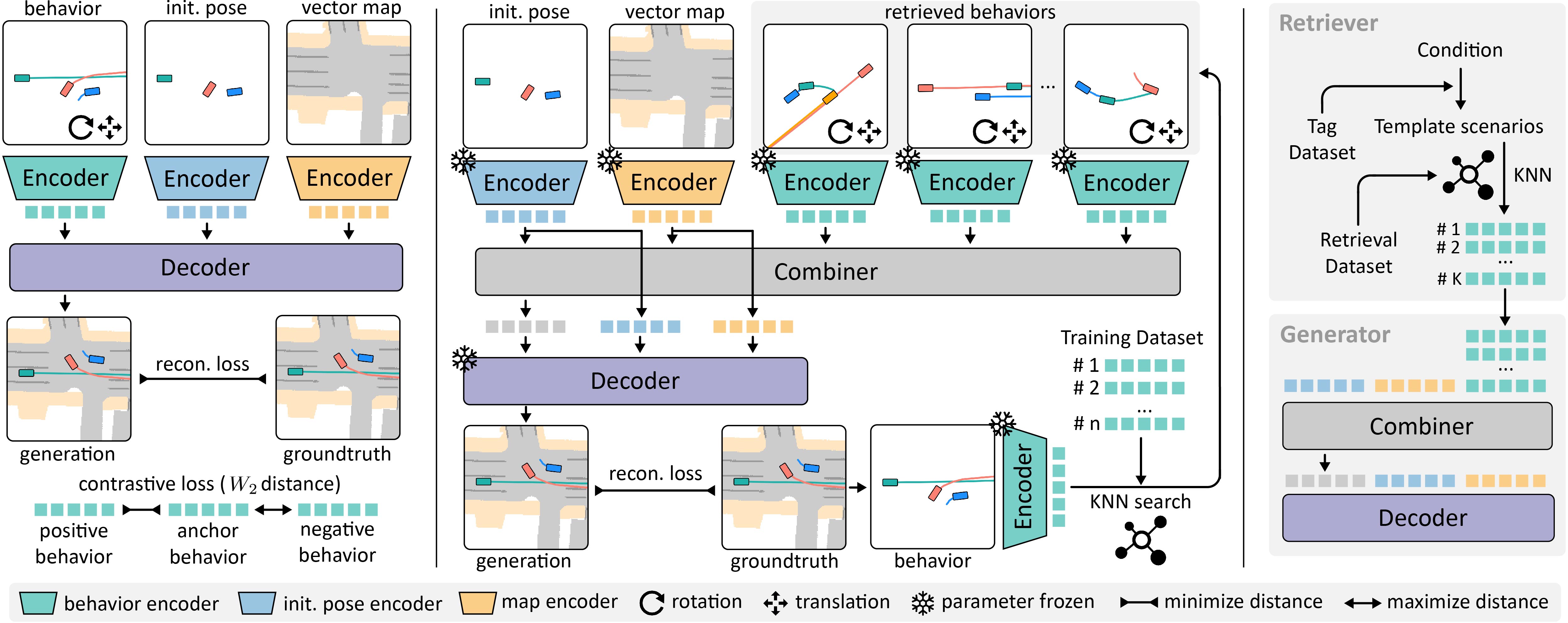}
    \vspace{-5mm}
    \caption{\textbf{Left}: the training pipeline for the encoders and decoder aimed at learning latent embeddings of scenarios. A contrastive loss is applied to behavior embeddings to ensure invariance to absolute positions. \textbf{Middle}: the training pipeline for the combiner with frozen encoder and decoder parameters. We use K-Nearest Neighbors (KNN) to retrieve scenarios similar to a template scenario in the dataset and use the retrieved behaviors to reconstruct the template scenario. \textbf{Right}: the generation pipeline with a retriever and a generator.}
    \vspace{-5mm}
    \label{fig:structure}
\end{figure*}

\subsection{Scenario Autoencoder}

Autoencoders~\cite{schmidhuber2015deep} are used to learn compressed latent representations of high-dimensional data, where an encoder projects the data into latent code and a decoder reconstructs the code in the data space. Given that traffic scenarios encompass spatial and temporal dynamics from multiple agents, we design a hierarchical encoder structure alternating between spatial and temporal layers based on transformer architecture~\cite{vaswani2017attention}. 
As depicted on the left of Figure~\ref{fig:structure}, our design incorporates three encoders for behavior, map, and initial position. 

First, we design a behavior encoder ${E}_{b}$ with a spatial-temporal transformer structure that comprises $L_{e}$ temporal transformer encoders ($\text{Encoder}_{t}^{i}$) and $L_{e}$ spatial transformer encoders ($\text{Encoder}_{s}^{i}$), where $i\in \{1,\dots,L_{e}\}$. 
The trajectory $\tau$ is first transformed into a latent embedding $z_b$, where $H$ denotes the hidden dimension, through a Multiple Linear Perception (MLP) module.
Subsequently, an alternating encoding procedure extracts spatial and temporal features from $z_b$.
To preserve temporal information, we add a sinusoidal positional embedding (PE) to the embedding before employing the temporal transformer.
To retain distinct agent information and further compress the behavior feature, we calculate the mean of the latent embedding across the temporal dimension, resulting in the final behavior embedding $z_b \in \mathbb{R}^{M\times H}$. 
Besides the behavior encoder, we also process the map information represented by lane vectors with the attention mechanism~\cite{vaswani2017attention}. 
Similarly to the first step in ${E}_{b}$, we project $m$ to the latent space with an MLP module.
Then we apply a multi-head attention module (MHA) with layer normalization (LN)~\cite{ba2016layer} to acquire the map embedding $z_m \in \mathbb{R}^{S\times H}$ with a learnable query embedding $q_m$. 

In the decoder, a spatial-temporal transformer architecture, similar to the behavior encoder, is designed to obtain the behavior embedding $z_b$ through $L_d$ times encoding.
Given that $z_b$ lacks the temporal dimension, we first replicate it $T$ times and then add a PE before entering it into the temporal transformer encoder. 
Throughout the decoding process, the map embedding $z_m$ is injected by a cross-attention mechanism:
$
    z_b \leftarrow z_b + \text{MHA}(z_b, z_m, z_m).
$
$\text{MHA}(Q, K, V)$ is the multi-head attention with $Q$, $K$, $V$ representing query, key, and value:
\begin{equation}
\begin{split}
    &\text{MHA}(Q, K, V) = \text{Concatenate}(h_1,...,h_i)W^O, \\
    &h_i = \text{Attention}(QW^Q_i, KW^K_i, VW^V_i),
\end{split}
\end{equation}
where $W^O$, $W^Q_i$, $W^K_i$, and $W^V_i$ are learnable parameters.
The last component of the decoder is an MLP that projects the hidden embedding back into the data domain, resulting in the reconstructed trajectory $\hat{\tau}$. 
To train the encoders and decoder, the mean square error is used as the reconstruction loss, expressed as $\mathcal{L}_{r} = | \hat{\tau} - \tau |_2$. 
Comprehensive descriptions of the encoding and decoding processes are provided in Algorithm~\ref{alg:encoder-decoder}.

It is important to note that, while the current design of the autoencoder is sufficient for acquiring compressed representations of scenarios, these representations are not invariant to the absolute coordinates and the order of the agents. This limitation can lead to substantial embedding distances between similar scenarios. To address these issues, two enhancements are proposed and detailed in the following section.

\SetKwProg{Behavior}{\textcolor{behavior-encoder}{\sf{\small{\textbf{Behavior Encoder}} $E_{b}$ ($\tau$):}}}{}{end}
\SetKwProg{Map}{\textcolor{map-encoder}{\sf{\small{\textbf{Map Encoder}} $E_{m}$ ($m$):}}}{}{end}
\SetKwProg{Init}{\textcolor{init-encoder}{\sf{\small{\textbf{Initial Pose Encoder}} $E_{i}$ ($\tau_0$):}}}{}{end}
\SetKwProg{Decoder}{\textcolor{decoder}{\sf{\small{\textbf{Decoder}} $D$ ($z_i, z_b, z_m$):}}}{}{end}
\newcommand\mycommfont[1]{\small\ttfamily\textcolor{gray}{#1}}
\SetCommentSty{mycommfont}
\begin{algorithm}[t]
\caption{{Details of Encoder and Decoder}}
\label{alg:encoder-decoder}
\begin{multicols}{2}
\footnotesize{
\Behavior{}{
    $z_b \leftarrow$ \textnormal{MLP ($\tau$)} \tcp{projection} 
    \For{$i$ \textnormal{in} $[1,...,L_e]$}{
        $z_b \leftarrow$ \textnormal{$\text{Encoder}_{t}^i$ (PE + $z_b$)} \\ 
        $z_b \leftarrow$ \textnormal{$\text{Encoder}_{s}^i$ ($z_b$)} \\
    }
    $z_b \leftarrow$ mean ($z_b$) \\ 
    \KwRet{\textnormal{behavior embedding} $z_b$} \\
}
\Map{}{
    Initialize a learnable query $q_m$ \\
    $z_m \leftarrow$ \textnormal{MLP ($m$)} \tcp{projection} 
    $z_m \leftarrow \textnormal{LN (MHA}\ (q_m, z_m, z_m))$ \\
    $z_m \leftarrow \textnormal{LN (} z_m + \textnormal{MLP}\ (z_m))$ \\
    \KwRet{\textnormal{map embedding} $z_m$} \\ 
}
\Init{}{
    $z_i \leftarrow$ MLP ($\tau_0$) \tcp{projection} 
    \KwRet{\textnormal{initial pose embedding} $z_i$} \\
}
\Decoder{}{
    $z_r \leftarrow z_b + \textnormal{MHA} (z_b, z_i, z_i)$ \\ 
    \For{$i$ \textnormal{in} $[1,...,L_d]$}{
        $z_r \leftarrow$ \textnormal{$\text{Encoder}_{t}^i$ (PE + $z_r$)} \\ 
        $z_r \leftarrow$ \textnormal{$\text{Encoder}_{s}^i$ ($z_r$)} \\
        $z_r \leftarrow z_r + \textnormal{MHA}^i (z_r, z_m, z_m)$ \\ 
    }
    $\hat{\tau} \leftarrow$ MLP ($z_r$) \tcp{projection} 
    \KwRet{\textnormal{reconstructed trajectory} $\hat{\tau}$} \\
}
}
\end{multicols}
\vspace{2mm}
\end{algorithm}

\subsection{Invariant feature with contrastive loss}

To enhance the representation of scenario similarity in the behavior embedding $z_b$, we employ contrastive learning to acquire invariant features. 
Specifically, we integrate InfoNCE~\cite{oord2018representation} as an additional loss function $\mathcal{L}_c$, which optimizes categorical cross-entropy to distinguish a positive sample from a batch of negative samples. 
In practice, for a query embedding $z_b$, a positive sample $z_b^+$ is generated by applying random rotation and translation to the original scenario ${\tau, m}$.
Meanwhile, negative samples $Z_b^-$ are selected from the remaining samples in the same batch. 
In conventional InfoNCE, to bring $z_b$ and $z_b^+$ closer together, the inner product is calculated between the embedding $z_b$ and the set ${ z_b^+, Z_b^- }$, employing cross-entropy loss to identify the index of the positive sample.

To ensure that the ordering of the agents does not affect the outcomes, $z_b \in \mathbb{R}^{M\times H}$ should exhibit permutation invariance across the dimension $M$. Otherwise, simply stacking them into a one-dimensional vector could erroneously represent distinct scenarios as significantly different. To establish permutation invariance, we adopt the Wasserstein distance $W_2$~\cite{villani2009optimal} as a more suitable metric than cosine distance for assessing the similarity. 
Taking $z_b$ as a distribution representing $M$ individual behaviors, the intuition behind this choice is that the Wasserstein distance identifies the minimal adjustments necessary for behaviors in one scenario to reflect those in another. With such a powerful tool, we have the following contrastive loss:
\begin{equation}
    \mathcal{L}_{c} = -\sum_{z_b}  \log \frac{\exp\left[ -W_2(z_b, z_b^{+}) \right]}{\sum_{z'\in \{z_b^+, Z_b^-\}}\exp \left[ -W_2(z_b, z')\right]}.
\end{equation}

The implementation of this loss results in the behavior embedding $z_b$ lacking absolute coordinate data, making the decoder incapable of accurately reconstructing the precise trajectory. To handle this problem, we add the initial poses $\tau_0$ of all agents as a supplementary input for the decoder. We encode these poses into $z_i$ with an MLP encoder $\text{Enc}_i$, and then integrate it into the decoder with an MHA module $z_r \leftarrow z_b + \text{MHA}(z_b, z_i, z_i)$. Additional specifics regarding the initial pose encoder are detailed in Algorithm~\ref{alg:encoder-decoder}.
In the training stage, the objective is to minimize a combined loss function $\mathcal{L} = \mathcal{L}_{r} + \lambda \mathcal{L}_{c}$, where $\lambda$ serves as a weighting factor for contrastive loss and is uniformly set at 0.1 for all experiments.

\section{Retrieval Augmented Scenario Generation}
\label{sec:retrieval}

The autoencoder structure introduced in Section~\ref{sec:embedding} builds a one-to-one generation framework, which still needs additional modules to support the retrieval augmented generation, which is a many-to-one framework.
In this section, we introduce a module called Combiner, which takes input behavior embeddings from multiple scenarios and outputs the combined embedding.
To train this module, we designed a KNN-based training pipeline that forces the model to learn to combine and edit existing scenarios. 

\subsection{The Training of Combiner}

Unlike the trajectory prediction task~\cite{shi2022motion} which has ground truth as the optimization target, the objective in training the combiner is to learn the alignment of behaviors in the retrieved scenarios with the initial pose and the specified map. A well-trained combiner should be able to compose behaviors from these retrieved scenarios, thereby generating new scenarios that have a resemblance to all the retrieved scenarios.
This type of objective aligns with the concept of meta-learning, or `learning to learn', as described in~\cite{hospedales2021meta}.

Within our training framework, for a given query scenario ${\tau, m}$ in the dataset, we initially utilize the behavior encoder to obtain $z_b$ and then use KNN to identify $K$ similar behavior embeddings from the database, denoted $z_{ret} = [z_{b,1},..., z_{b,K}]$.
Building on this, we propose a model comprising two MHA modules as the combiner:
\begin{equation}
\begin{split}
    & z_{rag} \leftarrow z_i + \text{MHA}(z_i, z_{ret}, z_{ret}), \\
    & z_{rag} \leftarrow z_{rag} + \text{MHA}(z_{rag}, z_m, z_m),
\end{split}
\label{eq:rag_map}
\end{equation}
where $z_i \leftarrow E_i(x_{0})$ and $z_m \leftarrow E_m(m)$ are the initial pose embedding and the map embedding of the query scenario.
Gradients are stopped for $z_i$ and $z_m$.
Assuming that the $K$ nearest scenarios adequately represent the query scenario, it should be feasible to reconstruct the behavior of the query scenario, $z_b$, using the retrieved scenario, $z_{rag}$. Consequently, we employ the following loss function as our training goal:
\begin{equation}
    \mathcal{L}_{rag} = \|D(z_i, z_{rag}, z_m) - \tau \|_2,
\end{equation}
wherein the parameters of the decoder $D\phi$ remain fixed during training. Essentially, this training method is designed to learn an "inverse" operation of the KNN, aiming to reconstruct a query scenario that closely resembles all $K$ retrieved scenarios in the database.

\subsection{The generation pipeline}

The pipeline for scenario generation, as delineated in Figure~\ref{fig:structure}, is categorized into retriever and generator components. The generation process is expedited by pre-processing all scenarios in the database using a behavior encoder, which yields behavior embeddings that facilitate efficient similarity computation.

The retriever enhances the versatility of generation by dividing the process into two stages. In the initial stage, users can choose a set of template scenarios that depict specific conditions. 
These may include manually annotated scenarios with tags denoting actions like left or right turns, enabling the generation of additional scenarios under similar tags or even a combination thereof. In addition, templates can include critical and interesting scenarios collected from real-world data. 
Solely relying on these templates for generation could be limited, which is mitigated by incorporating a secondary phase, which employs a KNN approach to fetch high-quality scenarios from a vast and unlabeled database to augment adaptability.

Subsequently, the generator, comprising a combiner and a decoder, follows the same inference as combiner training with the initial pose and lane map specified by the user. We obtain the RAG embedding $z_{rag}$ with Eq. (\ref{eq:rag_map}), and then infer the generated scenario through $\tau_{rag} \leftarrow D(z_i, z_{rag}, z_m)$. 

\section{Experiment}
\label{sec:experiment}

In this section, we start with an overview of the experimental setup and details regarding the implementation of the RealGEN framework.
Then we discuss the baselines, including different types of autoencoders and variants of RealGen, used for comparison. In the discussion of the findings, we respectively evaluate the quality of scenario embedding and the generation capability of RAG.

\subsection{Settings and Implementation Details}

We conducted all training and evaluation with the nuScenes~\cite{caesar2020nuscenes} dataset using the trajdata~\cite{ivanovic2023trajdata} package for data loading and processing. Each scenario spans a duration of 8 seconds, operating at a frequency of 2 Hz, and encompasses a maximum of 11 agents. We filter out agents that travel less than 3 meters in 8 seconds and select 11 agents closest to the ego vehicle. 
The map contains 100 lanes (each lane has 20 points) that are ordered according to the distance between the center of the lane and the center of the ego vehicle.

To train RealGen, we use Adam~\cite{kingma2014adam} as the optimizer to update the parameters of the autoencoder and the combiner.
To make the calculation of the Wasserstein distance efficient, we use the Sinkhorn distance~\cite{cuturi2013sinkhorn}, an entropy regularized approximation of the Wasserstein distance~\cite{villani2009optimal}, with implementation in the GeomLoss~\cite{feydy2019interpolating} package.

\subsection{Baselines}

In the experimental section, we evaluate the following reconstruction-based generative models as baselines for comparison.
{Autoencoder (\textbf{AE})} shares the same behavior encoder, map encoder, and decoder structures as RealGen, serving as the most straightforward baseline for scenario reconstruction.
\textbf{Contrastive AE} mirrors the structure of the RealGen autoencoder but omits the initial pose as absolute information.
\textbf{Masked AE} is a self-supervised learning baseline, which has been investigated for trajectory data as described in~\cite{wu2023masked, chen2023traj, yang2023rmp, cheng2023forecast}.
To evaluate the controllable generation capability, we select a state-of-the-art model\textbf{LCTGen}~\cite{tan2023language} as the baseline, which takes a high-level agent representation $z$ and a vector map $m$ as input. 
We also consider \textbf{LCTGen w/o $z$}, a variant model proposed in the original paper to show the reconstruction results without using the agent information.
Given that our method has two phases, we refer to the autoencoder component specifically as \textbf{RealGen-AE} and the complete model as \textbf{RealGen}.

\begin{figure*}[t]
    \centering
    \includegraphics[width=1.0\textwidth]{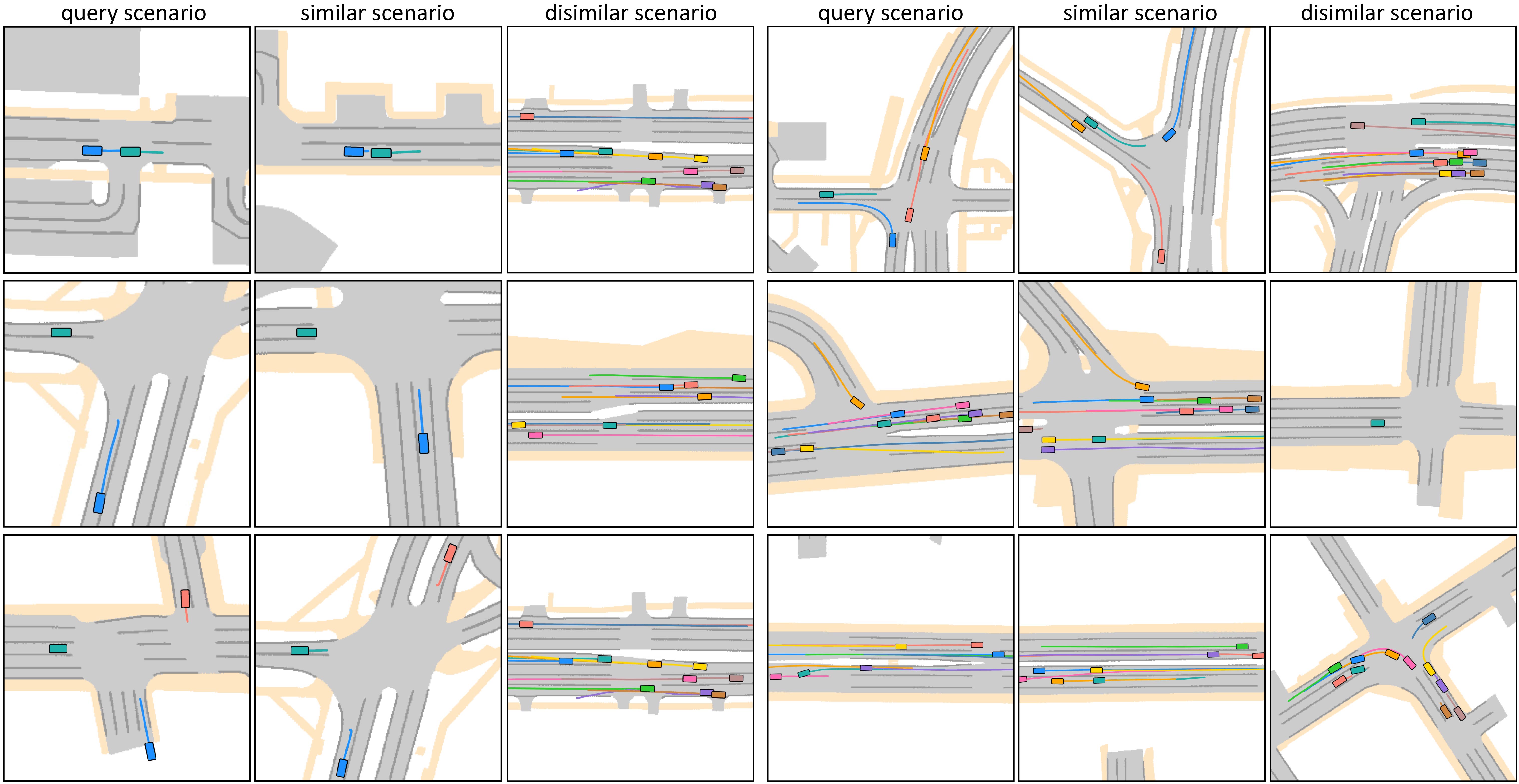}
    \vspace{-5mm}
    \caption{Qualitative evaluation of similar and dissimilar scenarios calculated by our scenario embedding. Rectangles represent the initial poses of vehicles and the lines represent the future trajectories.}
    \label{fig:similar_scene}
    \vspace{-3mm}
\end{figure*}

\subsection{Evaluation of Behavior Embedding}

We first evaluate the quality of the learned representation of behavior, which is critical for the following retrieval and generation processes.

\textbf{Visualizing similar and dissimilar scenarios.}
After training the auto-encoder with contrastive loss, the distance between behavior embeddings can be used as an indicator of the similarity of the two scenarios. To validate this statement, we visualize qualitative examples of using a query to find the most similar (minimal $W_2$ distance) and dissimilar (maximal $W_2$ distance) scenarios in Figure~\ref{fig:similar_scene}. We observe that the most similar scenario contains the same behavior and number of vehicles as in the query scenario.

\begin{figure}[t]
    \centering
    \includegraphics[width=0.75\textwidth]{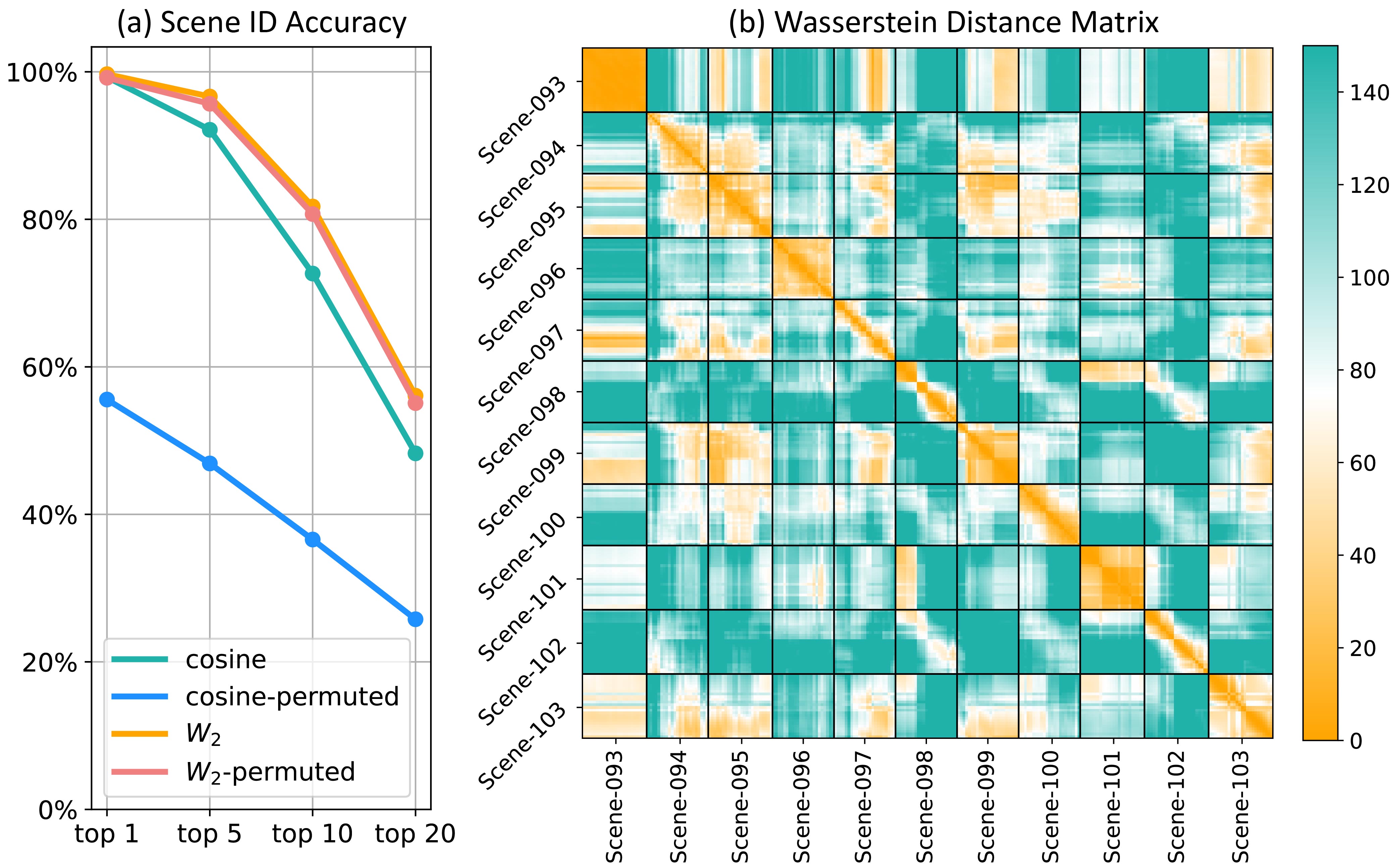}
    \vspace{-2mm}
    \caption{(a) Scene ID accuracy using the behavior embedding with difference distance metrics. (b) A matrix shows the Wasserstein distance between scenario segments, where each block contains the segments that belong to the same Scene ID.}
    \label{fig:scene_id}
    \vspace{-3mm}
\end{figure}

\textbf{Classifying Scene ID with behavior embedding.}
To further provide quantitative results on how well the behavior embedding encodes behavior information, we leverage the Scene ID of the nuScenes dataset. Each scene typically lasts 20 seconds and our scenario segments last 8 seconds, so we get multiple segments belonging to the same Scene ID. We assume that the segments having the same Scene ID have similar behaviors so that we can calculate the accuracy by using the distance to find the closest segment. In Figure~\ref{fig:scene_id}, we summarize the results in the left part, where top-$k$ means we calculate the accuracy for the closest $k$ segments. To validate the effectiveness of $W_2$ distance, we also permute the order of agents in the behavior embedding, which is denoted as \textit{cosine-permuted} and \textit{$W_2$-permuted}. We find that the cosine distance cannot deal with the permuted setting, as the order is important when stacking the embedding of agents. Our method performs well in top-1 and top-5 settings but badly in others, which could be explained by the fact that segments in one scene have very different behaviors. To validate this, we plot the distance matrix between segments for 11 scenes in the right part of Figure~\ref{fig:scene_id}. We find sub-blocks in the diagonal blocks, indicating that segments are similar in a small time interval but could be different when the time interval is large.

\begin{wraptable}{r}{0.5\textwidth}
\vspace{-12mm}
\caption{Accuracy of linear probing}
\label{tab:linear_probing}
\centering
\scriptsize{
\begin{tabular}{c|ccc}
    \toprule
    AE  &  Contrastive AE & Masked AE & RealGen-AE \\
    \midrule 
    82.5\% & 67.1\% & 86.2\%& \textbf{87.8\%} \\
    \bottomrule
\end{tabular}
}
\vspace{-7mm}
\end{wraptable}

\textbf{Linear probing of behavior embedding.}
Linear probing, a commonly employed method to assess representations in self-supervised learning (SSL), involves training a linear classifier using the derived embeddings~\cite{he2022masked}. To train this classifier, we implemented heuristic rules to assign basic behavioral labels (acceleration, deceleration, stopping, keeping speed, left/right turn) to each agent's embedding.
The outcomes, along with comparisons with baseline models, are presented in Table~\ref{tab:linear_probing}. These findings demonstrate that the embeddings generated by RealGen surpass all baseline models in terms of accuracy.

\subsection{Evaluation of Retrieval Augmented Generation}

Evaluating scene-level controllability and quality of scenario generation is an open problem due to the lack of quantitative metrics. Following most previous work~\cite{tan2023language, feng2023trafficgen, zhong2023guided}, we provide the results of the realism metrics and show the qualitative results of using RealGen for various downstream tasks. More results can be found in the appendix.

\begin{table*}[t]
\caption{Results of realism metrics. Recon.-based and Retrieval-based means using the target scenario and retrieved scenarios as input for generation, respectively.}
\vspace{-2mm}
\label{tab:realism}
\centering
\scriptsize{
\begin{tabular}{l|l|ccccccc}
    \toprule
    Category & {Method}  &  mADE & mFDE  & Speed   & Heading  & SCR & ORR \\
    \midrule 
    \multirow{4}{*}{\makecell{Recon.\\-based}} 
    & AE                   & \textbf{0.18\tiny{$\pm$0.03}}  & 0.41\tiny{$\pm$0.06} & \textbf{0.04\tiny{$\pm$0.01}} & 0.10\tiny{$\pm$0.01} & \textbf{0.02\tiny{$\pm$0.00}} & \textbf{0.02\tiny{$\pm$0.00}} \\
    & Masked AE            & 0.16\tiny{$\pm$0.01}  & \textbf{0.39\tiny{$\pm$0.01}} & \textbf{0.04\tiny{$\pm$0.01}} & \textbf{0.09\tiny{$\pm$0.01}} & 0.03\tiny{$\pm$0.00} & \textbf{0.02\tiny{$\pm$0.00}} \\
    & Contrastive AE       & 0.92\tiny{$\pm$0.02}  & 1.47\tiny{$\pm$0.04} & 0.12\tiny{$\pm$0.00} & 0.36\tiny{$\pm$0.02} & 0.04\tiny{$\pm$0.00} & 0.04\tiny{$\pm$0.00} \\
    & RealGen-AE           & 0.31\tiny{$\pm$0.01}  & 0.53\tiny{$\pm$0.01} & 0.08\tiny{$\pm$0.00} & 0.15\tiny{$\pm$0.01} & 0.03\tiny{$\pm$0.00} & \textbf{0.02\tiny{$\pm$0.00}} \\
    \midrule
    \multirow{5}{*}{\makecell{Retrieval\\-based}} 
    & AE-KNN               & 14.3\tiny{$\pm$0.03}  & 16.4\tiny{$\pm$0.05} & 0.57\tiny{$\pm$0.01} & 0.59\tiny{$\pm$0.02} & 0.15\tiny{$\pm$0.01} & 0.15\tiny{$\pm$0.01} \\
    & LCTGen               & 4.76\tiny{$\pm$0.09}  & 6.24\tiny{$\pm$0.08} & 0.52\tiny{$\pm$0.06} & 0.57\tiny{$\pm$0.03} & 0.07\tiny{$\pm$0.01} & 0.07\tiny{$\pm$0.01} \\
    & LCTGen w/o $z$       & 14.2\tiny{$\pm$0.07}  & 16.7\tiny{$\pm$0.09} & 2.04\tiny{$\pm$0.04} & 1.42\tiny{$\pm$0.00} & 0.16\tiny{$\pm$0.02} & 0.13\tiny{$\pm$0.04} \\
    & RealGen-AE-KNN       & 13.1\tiny{$\pm$0.06}  & 14.1\tiny{$\pm$0.03} & 0.46\tiny{$\pm$0.01} & 0.44\tiny{$\pm$0.00} & 0.12\tiny{$\pm$0.01} & 0.11\tiny{$\pm$0.00} \\
    & RealGen              & \textbf{1.54\tiny{$\pm$0.04}}  & \textbf{1.21\tiny{$\pm$0.03}} & \textbf{0.21\tiny{$\pm$0.03}} & \textbf{0.21\tiny{$\pm$0.01}} & \textbf{0.05\tiny{$\pm$0.00}} & \textbf{0.04\tiny{$\pm$0.00}} \\
    \bottomrule
\end{tabular}
}
\vspace{-5mm}
\end{table*}

\textbf{Realism of generated scenario.}
To evaluate the realism of generated scenarios, we consider the following metrics and show the results in Table~\ref{tab:realism}. We use the maximum mean discrepancy (MMD)~\cite{gretton2012kernel} to measure the similarity in velocity and direction between the original scenarios and the generated scenarios. We also compare the mean average displacement error (mADE) and the mean final displacement error (mFDE) for the average reconstruction performance. To evaluate scene-level realism, we calculate the scene collision rate (SCR) and the off-road rate (ORR) following the metrics defined in~\cite{tan2023language}.
The recon-based generation methods in Table~\ref{tab:realism} use the behavior of the target scenario as input. For this category, we compare RealGen-AE, only using the encoder and decoder modules, with three baseline methods. Due to the additional contrastive term, RealGen-AE achieves slightly worse performance than AE and Masked AE.
However, RealGen is designed for retrieval-based generation, which uses retrieved scenarios rather than the target scenario as input. To fairly compare the generation performance, we design two baselines named AE-KNN and RealGen-AE-KNN, which use KNN to find the most similar behavior embedding to the target scenario and use this embedding as input to the decoder for generation. According to the results, we find that RealGen achieves comparable performance as recon-based generation and outperforms baselines, which indicates the important role of the combiner in fusing the information of the retrieved scenarios.

\begin{figure*}[t]
    \centering
    \includegraphics[width=1.0\textwidth]{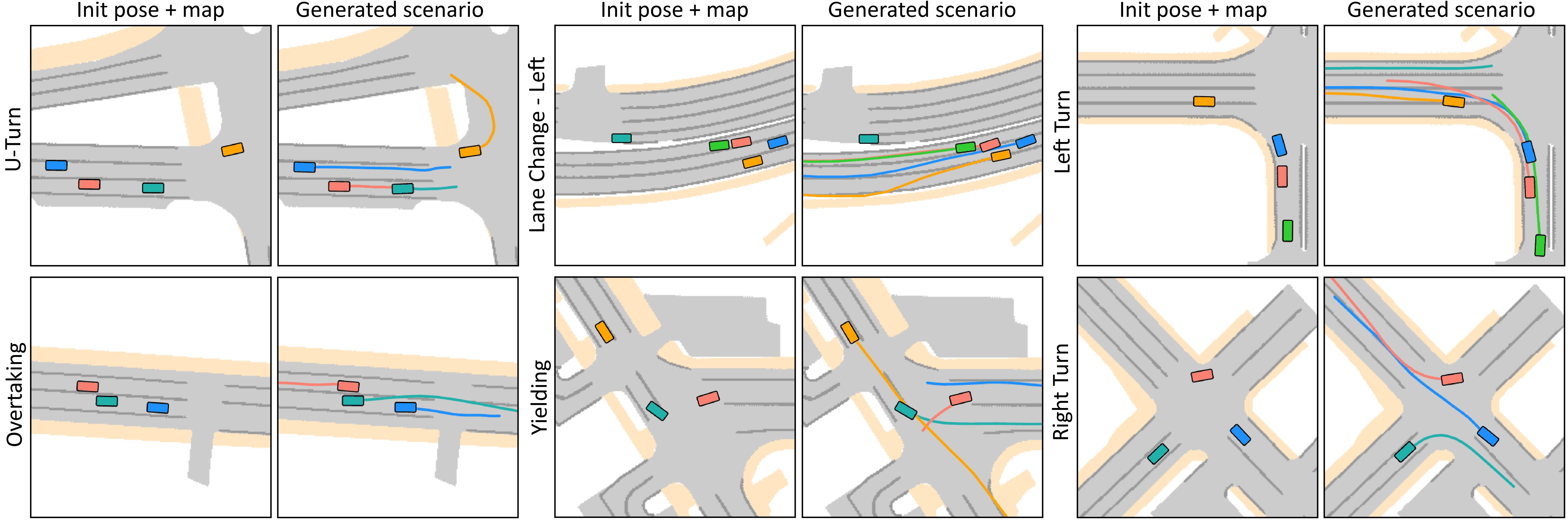}
    \vspace{-5mm}
    \caption{Examples of tag-retrieved scenarios generated by RealGen for six different tags.}
    \label{fig:scene_tag}
    \vspace{-2mm}
\end{figure*}

\begin{wrapfigure}{r}{0.5\textwidth}
    \centering
    \vspace{-7mm}
    \includegraphics[width=0.5\textwidth]{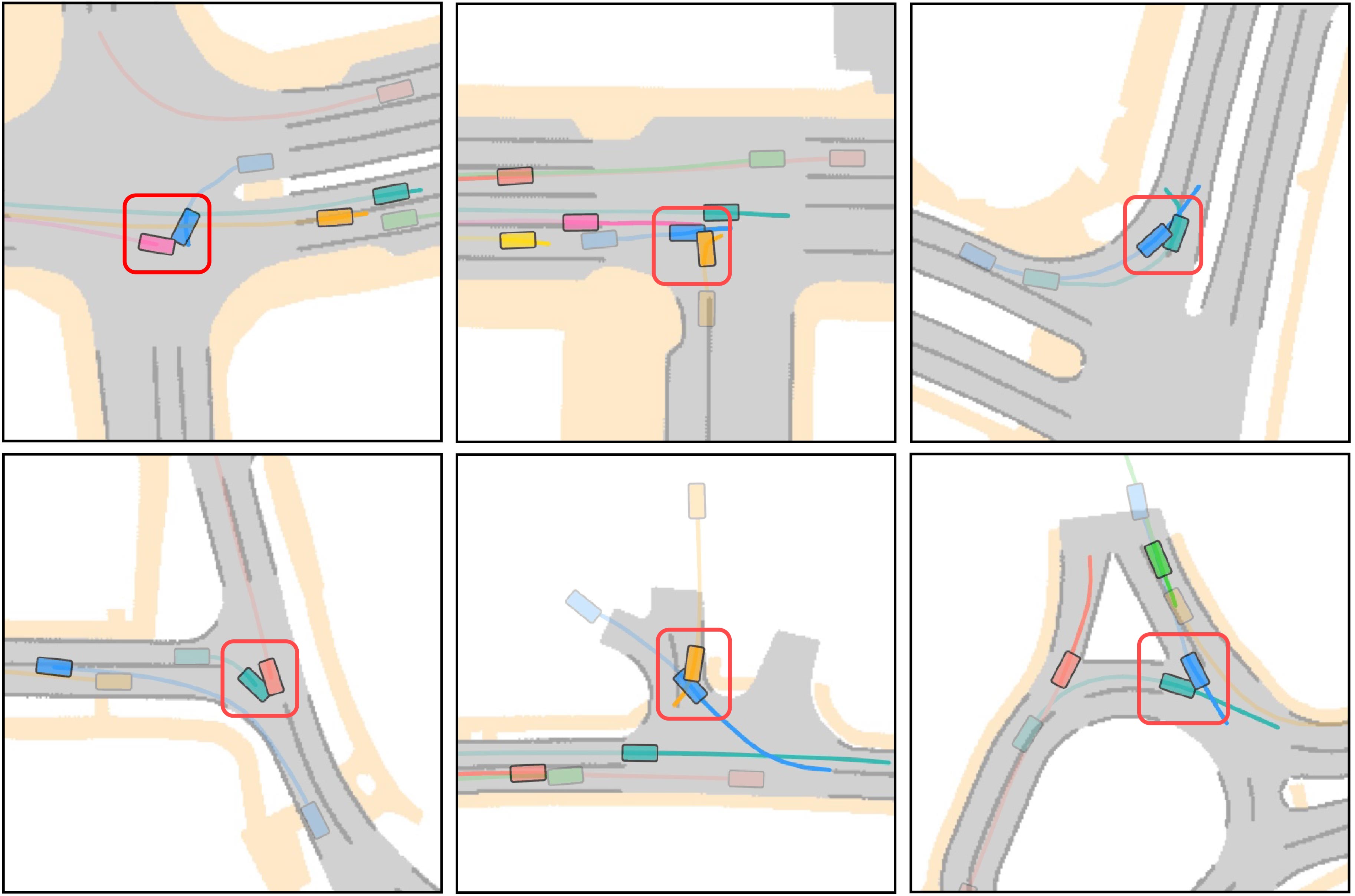}
    \vspace{-5mm}
    \caption{Examples of generating crash scenarios from RealGen. The shadow rectangles represent the initial positions of agents.}
    \label{fig:scene_crash}
    \vspace{-7mm}
\end{wrapfigure}

\textbf{Generating tag-retrieved scenarios.}
Now we explore the qualitative performance of using RealGen for tag-retrieved generation. Given a target behavior tag, we first obtain several template scenarios from a small dataset and use them to retrieve more scenarios from the training database, which will be used for generation in the combiner. Since there is no existing dataset with tags, we manually labeled six tags -- U-Turn, Overtaking, Left Lane Change, Right Lane Change, Left Turn, Right Turn -- for the nuScenes dataset to get template 1349 scenarios. We plot generated scenarios for each tag in Figure~\ref{fig:scene_tag}, where the left part of each example shows the given initial pose and map and the right part of each example shows the generated scenario from RealGen.  

\begin{wraptable}{r}{0.55\textwidth}
\vspace{-11mm}
\caption{Safety-critical scenario generation.}
\label{tab:collision}
\centering
\scriptsize{
\begin{tabular}{c|ccc}
    \toprule
    Method           & Realgen-AE-R & RealGen-R & RealGen \\
    \midrule 
    Collision Rate ($\uparrow$)  & 0.92 & 0.83 & \textbf{0.59} \\
    \bottomrule
\end{tabular}
}
\vspace{-7mm}
\end{wraptable}

\textbf{Safety-critical scenario generation.}
Beyond the tag mentioned above, RealGen demonstrates its capacity for in-context learning by generating critical and unseen crash scenarios, divergent from those of the training datasets. This is initiated by manually creating several crash scenario templates, guided by the scenarios recorded in the NHTSA Crash Report~\cite{nhtsacrash}. Subsequently, we generate crash scenarios using existing initial poses and maps of the dataset. Figure~\ref{fig:scene_crash} illustrates six instances in which the shadowed rectangles denote the initial positions of agents and the red box highlights the point of collision.
To quantitatively assess the performance of safety-critical scenarios, we compare RealGen with two baselines. \textit{RealGen-AE-R} random samples behavior embeddings in the Realgen-AE model, and \textit{RealGen-R} random retrieves behavior embeddings. According to the results in Table~\ref{tab:collision}, we find that the scenarios generated by RealGen using crash scenarios as templates achieve the highest collision rate, which means that RealGen has more efficiency.

\textbf{Human evaluation of controllability.} As there is no automatic way to evaluate controllability, we follow the protocol in~\cite{tan2023language} to perform A/B testing using human evaluation. We report the ratio of our method preferred as well as the absolute score (0-5) of both our method and the baseline. In Table~\ref{tab:controllability}, we find that scenarios generated by RealGen are highly preferred in most categories. The absolute score of RealGen is also much higher than that of LCTGen.

\begin{table*}[t]
\caption{Results of human evaluation of controllability. (Details in Appendix~\ref{app:human})}
\vspace{-2mm}
\label{tab:controllability}
\centering
\scriptsize{
\begin{tabular}{l|c|c|c|c|c}
    \toprule
    Category                 & {Left-Turn}  & {Right-Turn} & {Left-Lane-Change}  & {Right-Lane-Change}   & {Straight}  \\
    \midrule 
    RealGen Preferred ($\%$) & 81.8          & 91.7          & 97.8         & 93.3 & 100.0 \\
    RealGen Score (0-5)      & \textbf{4.27$\pm$1.05} & \textbf{4.27$\pm$0.69} & \textbf{3.96$\pm$1.94} & \textbf{4.17$\pm$1.93} & \textbf{3.94$\pm$2.27} \\
    LCTGen Score (0-5)       & 2.15$\pm$1.44 & 2.08$\pm$1.19 & 2.42$\pm$1.95 & 2.0$\pm$1.96 & 2.14$\pm$2.31 \\
    \bottomrule
\end{tabular}
}
\vspace{-5mm}
\end{table*}

\begin{wraptable}{r}{0.57\textwidth}
\vspace{-11mm}
\caption{Downstream task evaluation.}
\label{tab:downstream}
\centering
\scriptsize{
\begin{tabular}{c|ccc}
    \toprule
    Method                         & Original & Random Aug. & RealGen Aug. \\
    \midrule 
    mADE ($\downarrow$)            & 3.544    & 2.920       & \textbf{2.309} \\
    collision rate ($\downarrow$)  & 0.049    & 0.037       & \textbf{0.018} \\
    \bottomrule
\end{tabular}
}
\vspace{-8mm}
\end{wraptable}

\textbf{Downstream task evaluation.}
A direct downstream task of our method is to use the generated data to augment the training dataset of trajectory prediction models. 
We use Autobots~\cite{girgis2021latent} as a predictor and report the results trained on different datasets in Table~\ref{tab:downstream}. \textit{Original} means using the original data in nuScenes~\cite{caesar2020nuscenes}, \textit{Random Aug.} means augmenting the original dataset with Gaussian noise, and \textit{RealGen Aug.} means augmenting the original dataset with scenarios from RealGen. We observe that the model trained with the RealGen dataset achieves the lowest mADE and collision rate.


\vspace{-3mm}
\section{Conclusion and Limitation}
\vspace{-2mm}
\label{sec:conclusion}

This paper proposes RealGen, a novel framework for traffic scenario generation that utilizes retrieval-augmented generation. 
Unlike previous approaches, which primarily rely on models replicating training distributions, RealGen demonstrates in-context learning abilities that synthesize scenarios by combining and modifying provided examples, enabling controlled generation.
These scenarios can be automatically obtained from a retrieval system, which only requires the users to provide a few template scenarios as examples. We comprehensively evaluated the similarity autoencoder model for retrieval and the combiner model for generation. The findings indicate that RealGen achieves low reconstruction error and high generation quality.

The primary limitation of our current method is that the behavior encoder focuses solely on agent trajectories, neglecting the intricate interactions between agents and lane maps in complex behaviors. Advancing the feature representation within the behavior encoder could significantly broaden RealGen's capacity to generate diverse and controllable traffic scenarios.


\section*{Acknowledgments}
Wenhao Ding contributed to this paper while being an intern at NVIDIA Research. Ding Zhao was partially supported by the National Science Foundation under grant CNS-2047454.

%
%
\bibliographystyle{eccv2024/splncs04}
\bibliography{main}

\end{document}